\documentclass{article}
\usepackage{graphicx}
\usepackage{mathrsfs} 
\usepackage{subfig}  
\usepackage[subfigure]{tocloft} 
\usepackage{fancyhdr}
\usepackage[english]{babel}
\usepackage{csquotes}
\usepackage{footmisc}
\usepackage{algorithmic}
\usepackage{listings}
\usepackage{array}     
\usepackage{booktabs} 
\usepackage{fancyvrb}
\usepackage{enumitem}  
\usepackage{mathtools} 
\usepackage{multirow}  
\usepackage{amsmath,amssymb} 
\usepackage{scrextend} 
\usepackage[]{acronym} 
\usepackage{esint}
\usepackage{mathrsfs} 
\usepackage{authblk}
\usepackage[left=3cm,top=3cm,right=3cm,bottom=3cm,bindingoffset=0.5cm]{geometry}
\usepackage[title]{appendix}
\usepackage{geometry}
\usepackage{amsmath, amssymb, bm, mathtools}
\usepackage{enumitem}

\usepackage[utf8]{inputenc}
\usepackage{amsmath}
\usepackage{amsfonts}
\usepackage{graphicx}
\usepackage{listings}
\usepackage{xcolor}

\makeatother
\newcommand{\tenq}[1]{\hbox{\oalign{$\bm{1}$\crcr\hidewidth$\scriptscriptstyle\bm{\approx}$\hidewidth}}}   
\DeclareMathSymbol{\ii}{\mathalpha}{letters}{"10}
\DeclareMathSymbol{\jj}{\mathalpha}{letters}{"11}

\newcommand\norm[1]{\left\lVert1\right\rVert} 
\newcommand{\mynorm}[1]{ \left | 1 \right | } 
\newcommand{\quotes}[1]{``#1''}  

\title{Long-Sequence Memory with Temporal Kernels and Dense Hopfield Functionals}
\author{A. Farooq}
\date{} 

\affil[]{University of New Brunswick}
\begin{document}
\maketitle

\begin{abstract}
In this study we introduce a novel energy functional for long-sequence memory, building upon the framework of dense Hopfield networks which achieves exponential storage capacity through higher-order interactions. Building upon earlier work on long-sequence Hopfield memory models \cite{chaudhry2024long}, we propose a temporal kernal $K(m, k)$ to incorporate temporal dependencies, enabling efficient sequential retrieval of patterns over extended sequences. We demonstrate the successful application of this technique for the storage and sequential retrieval of movies frames which are well suited for this because of the high dimensional vectors that make up each frame creating enough variation between even sequential frames in the high dimensional space. The technique has applications  in modern transformer architectures, including efficient long-sequence modeling, memory augmentation, improved attention with temporal bias, and enhanced handling of long-term dependencies in time-series data. Our model offers a promising approach to address the limitations of transformers in long-context tasks, with potential implications for natural language processing, forecasting, and beyond.  
\end{abstract}

\section{Introduction}

Hopfield networks, introduced by John Hopfield in 1982 \cite{hopfield1982neural}, are a foundational model of associative memory in neural networks. These classical networks store binary patterns as stable states within an energy function:

\begin{equation}
E = -\frac{1}{2} \sum_{i,j} w_{ij} s_i s_j
\end{equation}

where $ s_i \in \{-1, 1\} $ represents the state of neuron $ i $, and $ w_{ij} $ is the synaptic weight between neurons $ i $ and $ j $. Despite their elegance, classical Hopfield networks suffer from a storage capacity that scales linearly with the number of neurons and are sensitive to correlated patterns, limiting their applicability to complex sequential tasks.\\\

In 2016, Krotov and Hopfield \cite{krotov2016dense} introduced {dense Hopfield networks} , incorporating higher-order interactions into the energy function:

\begin{equation}
E = - \sum_{p=2}^{P} \frac{1}{p!} \sum_{i_1, \ldots, i_p} J_{i_1 \ldots i_p} s_{i_1} \cdots s_{i_p}
\end{equation}

where $ J_{i_1 \ldots i_p} $ are higher-order weights, and $ p $ is the interaction order. This formulation achieves an exponential storage capacity, enabling the network to store complex patterns. In a follow-up work, Krotov and Hopfield \cite{krotov2019dense} demonstrated that dense Hopfield networks are robust to adversarial inputs, enhancing their suitability for sequential memory tasks where noise is prevalent.\\

Building on this, Demircigil et al. \cite{demircigil2017model} further advanced the theoretical foundation by showing that an exponential interaction function in Hopfield networks yields a storage capacity of up to $ 2^{\Theta(d)} $ patterns in $ \mathbb{R}^d $. Their energy function incorporates an exponential term, 

\begin{equation}
 E = - \sum_{i} \exp(\beta \sum_{j} w_{ij} s_i s_j)
\end{equation} 

providing a mathematical basis for the high-capacity memory models that followed.\\

The connection between Hopfield networks and attention mechanisms was solidified by Ramsauer et al. \cite{ramsauer2021hopfield} who proposed modern Hopfield networks with continuous states, defined by the energy function:

\begin{equation}
E = - \frac{1}{\beta} \log \left( \sum_{k=1}^N \exp(\beta \langle \mathbf{s}, \mathbf{\xi}_k \rangle) \right) + \frac{\lambda}{2} \|\mathbf{s}\|^2
\end{equation}

where $ \mathbf{s} \in \mathbb{R}^d $ is the state vector, $ \mathbf{\xi}_k $ are stored patterns, $ \beta $ controls sharpness, and $ \lambda $ is a regularization parameter. A key insight is that the update rule derived from this energy function:

\begin{equation}
\mathbf{s}_{\text{new}} = \sum_{k=1}^N \text{softmax}(\beta \langle \mathbf{s}, \mathbf{\xi}_k \rangle) \mathbf{\xi}_k
\end{equation}

is equivalent to the attention mechanism in transformers \cite{vaswani2017attention}, bridging associative memory with deep learning. This equivalence highlights the potential of Hopfield networks to enhance transformer architectures for tasks requiring long-context understanding.\\

Parallel efforts in sequenced memory have also contributed to this landscape. Sukhbaatar et al. \cite{sukhbaatar2015end} introduced \textit{end-to-end memory networks} in 2015, a precursor to modern memory-augmented architectures. Their model uses a soft attention mechanism over memory slots, $ \text{output} = \sum_{i} p_i \mathbf{m}_i $, where $ p_i = \text{softmax}(\mathbf{q}^T \mathbf{m}_i) $, to retrieve relevant information for question answering tasks.\\

Ba et al. \cite{ba2016using} proposed \textit{fast weights} to store recent memories in neural networks, using an update rule $ W_t = \lambda W_{t-1} + \eta \mathbf{x}_t \mathbf{y}_t^T $, where $ \lambda $ controls the decay of past information, offering a temporal weighting mechanism for sequential data.\\

More recently, Tran and Yanushkevich \cite{tran2023hopfield} explored the integration of Hopfield networks with transformers for long-sequence processing in NLP. They introduced a hybrid model with a Hopfield-inspired energy function incorporating temporal decay, 

\begin{equation}
E = - \sum_{k} \alpha^{t-k} \exp(\beta \langle \mathbf{s}, \mathbf{\xi}_k \rangle) 
\end{equation}

demonstrating improved performance on tasks requiring extended context, such as document-level question answering.\\

The challenge of processing long sequential data in Hopfield networks was directly addressed by Chaudhry, Krotov, Pahlevan et al. \cite{chaudhry2024long} in their work on long-sequence Hopfield memory where they extended dense Hopfield networks to handle extended time horizons, likely through mechanisms that incorporate temporal dynamics or memory slots, enabling the retention of context in applications such as language modeling and time-series analysis. Their key contribution was the progression of the dynamic evolution equation, deriving a continuous-time update rule of the form

\begin{equation}
\frac{d \mathbf{s}}{dt} = - \nabla_{\mathbf{s}} E(\mathbf{s}, t) + \gamma \sum_{k} \alpha^{t-k} \mathbf{s}^{(k)}
\end{equation}

where $ \gamma $ modulates the influence of past patterns $ \mathbf{s}^{(k)} $ with a decay factor $ \alpha $, enabling the network to retrieve patterns over extended sequences while preserving the exponential storage capacity established by Krotov et al. \cite{krotov2016dense} and Demircigil et al. \cite{demircigil2017model}. However, their approach relied on a fixed temporal decay, limiting adaptability to varying sequence dynamics. \\

In this paper, we build upon these advancements by introducing a modified energy functional that integrates temporal kernels to enhance the handling of long-sequence data. Our innovation introduces a kernel offering adaptive temporal weighting and enhanced stability through regularization. Drawing inspiration from the deep learning relevance of Ramsauer et al. \cite{ramsauer2021hopfield}, temporal weighting in fast weights \cite{ba2016using}, and hybrid Hopfield-transformer models \cite{tran2023hopfield}, our unified framework,  promises significant advancements in transformer architectures for long-sequence tasks, with potential applications in video processing, NLP, and time-series analysis.\\

The rest of this paper is structured as follows: Section 2 details the formulation of our sequenced energy functional, and Section 3 presents te form of this functional suitable for movie frame retrieval.  Setion 4 has results of numerical experiments, and Section 5 discusses the conclusions and future directions.

\section{A New Dense Energy Functional with Temporal Kernels}

In this section, we introduce a general energy functional designed for sequential pattern retrieval, leveraging a temporal kernel $ K $ and a functional $ F $ inspired by dense Hopfield networks. This formulation enables the modeling of long-sequence memory, with applications to modern sequence-processing architectures such as transformers. We first present the general framework, then specialize the kernel $ K $ to a Gaussian form and discuss its advantages, and finally explore two distinct choices for $ F $: an exponential form proposed by Demircigil et al. and a log-sum-exp form due to Ramsauer et al. \cite{ramsauer2021hopfield}.\\

Consider a system with a current state $ \mathbf{s} \in \mathbb{R}^d $ and a set of stored patterns $ \mathbf{s}^{(k)} \in \mathbb{R}^d $, for $ k = 0, \ldots, N-1 $, where each pattern is normalized such that $ \langle \mathbf{s}^{(k)}, \mathbf{s}^{(k)} \rangle = d $. At a discrete time step $ m $, we define the energy functional as:

\begin{equation}
E(m, \mathbf{s}) = \sum_{k=0}^{N-1} K(m, k) F(\beta \langle \mathbf{s}, \mathbf{s}^{(k)} \rangle) + \frac{\lambda}{2} \|\mathbf{s}\|^2
\end{equation}

where $ K(m, k) \geq 0 $ is a temporal kernel function that weights the influence of the $ k $-th stored pattern $ \mathbf{s}^{(k)} $ based on its temporal distance from the current time step $ m $,  $ F: \mathbb{R} \to \mathbb{R} $ is a functional that shapes the energy landscape, determining the interaction between the current state and stored patterns. $ \beta > 0 $ is a sharpness parameter that controls the steepness of $ F $ and $ \frac{\lambda}{2} \|\mathbf{s}\|^2 $ is a regularization term with $ \lambda > 0 $, ensuring stability of the state $ \mathbf{s} $.\\

The kernel $K(m, k)$ modulates the contribution of each pattern according to its relevance at time $m$, facilitating sequential retrieval by emphasizing temporally proximate patterns. The functional $F$ defines the energy wells around each stored pattern, and its specific form significantly impacts the system’s storage capacity and dynamics.  Let us consider below two simple forms of $F$ and finally a Gaussian form for the kernal function $K(m,k)$.

\subsection{Exponential from of the Functional $F$}

We now explore a specific choice for $ F $, inspired by the work of Demircigil et al. \cite{demircigil2017model}, who demonstrated that an exponential interaction function yields exponential storage capacity in dense Hopfield networks. We define:

\begin{equation}
F(x) = - \exp(x)
\end{equation}

so that the energy functional becomes:

\begin{equation}
E(m, \mathbf{s}) = - \sum_{k=0}^{N-1} K(m, k) \exp(\beta \langle \mathbf{s}, \mathbf{s}^{(k)} \rangle) + \frac{\lambda}{2} \|\mathbf{s}\|^2.
\end{equation}

To understand the dynamics, we compute the gradient with respect to $\mathbf{s}$:

\begin{equation}
\nabla_{\mathbf{s}} E(m, \mathbf{s}) = - \beta \sum_{k=0}^{N-1} K(m, k) \exp(\beta \langle \mathbf{s}, \mathbf{s}^{(k)} \rangle) \mathbf{s}^{(k)} + \lambda \mathbf{s}
\end{equation}

This gradient drives the continuous-time update rule:

\begin{equation}
\frac{d \mathbf{s}}{dt} = - \nabla_{\mathbf{s}} E(m, \mathbf{s}) = \beta \sum_{k=0}^{N-1} K(m, k) \exp(\beta \langle \mathbf{s}, \mathbf{s}^{(k)} \rangle) \mathbf{s}^{(k)} - \lambda \mathbf{s}
\end{equation}

This update resembles a weighted sum of the stored patterns, where the weights $K(m, k) \exp(\beta \langle \mathbf{s}, \mathbf{s}^{(k)} \rangle)$ combine temporal proximity (via $ K $) and pattern similarity (via the exponential term). The exponential form amplifies the contribution of patterns with high similarity to $ \mathbf{s} $, creating deep energy wells that enhance retrieval accuracy.\\

The key advantage of this choice, as shown by Demircigil et al. \cite{demircigil2017model}, is its capacity to store an exponential number of patterns—up to $ 2^{\Theta(d)} $ in $ \mathbb{R}^d $—far exceeding the polynomial limits of classical Hopfield networks. This makes it particularly suitable for applications requiring the retention of vast sequences, such as language modeling or time-series prediction.

\subsection{Log-Sum-Exp (LSE) Functional}

An alternative choice for $ F $ is the log-sum-exp functional, proposed by Ramsauver et al. \cite{ramsauer2021hopfield}, which provides a smooth approximation to the maximum function and connects directly to attention mechanisms in transformers. We adapt it to our framework as follows:

\begin{equation}
E(m, \mathbf{s}) = - \frac{1}{\beta} \log \left( \sum_{k=0}^{N-1} K(m, k) \exp(\beta \langle \mathbf{s}, \mathbf{s}^{(k)} \rangle) \right) + \frac{\lambda}{2} \|\mathbf{s}\|^2.
\end{equation}

Here, the term $ - \frac{1}{\beta} \log \left( \sum_{k} K(m, k) \exp(\beta \langle \mathbf{s}, \mathbf{s}^{(k)} \rangle) \right) $ replaces the sum in the general formulation, effectively acting as a single energy term that aggregates contributions from all patterns.\\

The gradient is:

\begin{equation}
\nabla_{\mathbf{s}} E(m, \mathbf{s}) = - \frac{\sum_{k=0}^{N-1} K(m, k) \exp(\beta \langle \mathbf{s}, \mathbf{s}^{(k)} \rangle) \mathbf{s}^{(k)}}{\sum_{k=0}^{N-1} K(m, k) \exp(\beta \langle \mathbf{s}, \mathbf{s}^{(k)} \rangle)} + \lambda \mathbf{s},
\end{equation}

where the first term is a normalized weighted average of the patterns $ \mathbf{s}^{(k)} $, with weights $ K(m, k) \exp(\beta \langle \mathbf{s}, \mathbf{s}^{(k)} \rangle) $ forming a softmax-like distribution over the patterns.\\

This functional, like the exponential as high storage capacity.  It also has the advantage that 
its gradient form mirrors the attention mechanism in transformers, where the state is updated based on a weighted combination of stored patterns (or keys), weighted by a softmax over similarity scores.  This choice is particularly advantageous in contexts where the system must balance contributions from multiple patterns, such as in attention-based models processing long sequences, offering a robust alternative to the exponential functional.\\

\subsection{Gaussian Temporal Kernel}

A natural and effective choice for the temporal kernel is the Gaussian function, defined as:

\begin{equation}
K(m, k) = \tilde{w}_k(m) = \exp\left(-\frac{(m - k)^2}{2 \sigma^2}\right)
\end{equation}

where $ \sigma > 0 $ is a parameter that controls the width of the Gaussian. This kernel assigns higher weights to patterns $ \mathbf{s}^{(k)} $ whose time indices $ k $ are closer to the current time step $ m $, with the influence decaying smoothly as the temporal distance $ |m - k| $ increases.\\

The Gaussian kernel offers several key advantages: (i) Smooth Temporal Weighting: The exponential decay ensures that patterns near $ m $ contribute more significantly to the energy, while distant patterns have a diminished but non-zero influence, providing a smooth transition across the sequence. (ii)  Adjustable Temporal Focus: The parameter $ \sigma $ allows precise control over the temporal scope. A small $ \sigma $ concentrates the system’s attention on a narrow window around $ m $, while a larger $ \sigma $ broadens the context, incorporating more patterns into the energy computation. (iii) Limiting Behavior: As $ \sigma \to 0 $, the Gaussian approximates a Dirac delta function, $ K(m, k) \to \delta_{m k} $, reducing the energy to focus solely on the pattern at $ k = m $. This enables strict sequential retrieval when desired.\\

These properties make the Gaussian kernel highly versatile, balancing local precision and global context, which is critical for tasks involving sequential memory and long-range dependencies.
    
In the next section to adopt the log-sum-exp form of $F$ with some added regularization terms for the purpose of sequential storage and retrieval of movie frames.

\section{Sequential Movie Frame Retrieval}

We wish to develop an energy functional $E$ such that $N$ movie frames given by $N$ state vectors $\mathbf{s}^{(0)}, \mathbf{s}^{(1)}, \ldots, \mathbf{s}^{(N-1)}$,  where $ \mathbf{s}^i \in \mathbb{R}^d $, are encoded as preferred states (e.g., minima or stable points) in a Modern Hopfield network. Minimizing $E$ sequentially reveals these frames in the correct order, like stepping through attractors in a Hopfield-like network. The process mimics \quotes{playing} the movie by traversing an energy landscape designed to reflect the sequence. \\

Sequential retrieval of movie frames aligns naturally with the temporal structure of our energy functional. Movie frames are inherently ordered in a temporal sequence, where each frame \( \mathbf{s}^{(k)} \in \mathbb{R}^d \) at time index \( k \) represents a visual snapshot that evolves smoothly or transitions abruptly (e.g., scene changes). The Gaussian kernel \( K(m, k) \) is designed to weight patterns based on their temporal proximity to the current time step \( m \), assigning higher influence to nearby frames while allowing contributions from more distant frames to decay smoothly. This property mirrors human perception of video continuity, where recent frames are more relevant for predicting or recalling the next frame, but recurring motifs (e.g., a character reappearing) from earlier frames can still contribute to the context. For instance, setting \( \sigma = 10 \) might emphasize frames within a few seconds of the current frame, while a larger \( \sigma \) could capture longer-term dependencies across minutes of footage.\\

We will be working with discrete time defined as: $t \in [t_0,t_1, t_2, \ldots, t_{N-1} ]$, where $t_0$, $t_1$, etc. are discrete time stamps at equispaced intervals.  At $t_0$ the state vector is denoted as $\mathbf{s}^{(0)}$, and at $t_1$ it is $\mathbf{s}^{(1)}$, and so on. We will set $t_0 = 0$ and $t_i = (i-1)\Delta t$, where $\Delta t = T/(N-1)$, $T$ being the total elapsed time and $N-1$ being the number of intervals. Without loss of generality, we will be that assuming $\Delta t = 1 $, which gives $t \in [0,1, 2, \ldots, m, \ldots {N-1} ]$.\\

We propose the following continuous state, time dependent energy functional $E(\mathbf{s},m)$ inspired by the approach of Ramsauer et al. \cite{ramsauer2021hopfield}:

\begin{equation}
E(\mathbf{s}, m) = \frac{\lambda}{2} \|\mathbf{s}\|^2 + \lambda_f \left\| \mathbf{s} - \mathbf{s}^{(m)} \right\|^2 + \mu \left\| \mathbf{s} - \mathbf{s}^{(m-1)} \right\|^2 - \frac{1}{\beta} \log \left( \sum_{k=0}^{N-1} w_k(m) e^{\beta \langle \mathbf{s}, \mathbf{s}^{(k)} \rangle} \right) - \max_k \langle \mathbf{s}, \mathbf{s}^{(k)} \rangle
\label{eqn:discrete-enhanced}
\end{equation}

where $\mathbf{s} \in \mathbb{R}^d$ is the state vector, $\mathbf{s}^{(k)} \in \mathbb{R}^d$ are $N$ stored patterns, $\langle \mathbf{s}, \mathbf{s}^{(k)} \rangle$ is the inner product, $\beta > 0$ controls the sharpness of the log-sum-exp (LSE) term, $\lambda > 0$ is a regularization parameter, $\lambda_f > 0$ is a fidelity parameter anchoring $\mathbf{s}$ to the target frame $\mathbf{s}^{(m)}$ at time $t=m$. $ \mu \geq 0 $ is a continuity parameter penalizing deviations from the previously retrieved frame. $ w_k(m) $ are time-dependent Gaussian weights as defined below\\

The energy surface given by $E(\mathbf{s},m)$ is designed so that seeking its minimum will lead to $\mathbf{s}^{(m)}$.  \\

The above functional is designed to provide a mechanism for robust  \emph{sequential} frame retrieval in our Modern Hopfield Network, to ensure reliable convergence to the target frame $ \mathbf{s}^{(m)} $ at each time step $ t=m $, while promoting smooth transitions between consecutive frames to mimic movie-like playback. \\

It may be noted that the energy functional, apart from the temporal weight kernels includes a continuity term, a fidelity term, and enforces normalization on both the time-dependent weights and the stored frames. Below, we present the mathematical formulation of these terms, along with their justifications.\\

\subsection{Definition of Weights}

In general there can be many choices for $w_k(m)$. We have found the Gaussian kernal, 

\[
\tilde{w}_k(m) = \exp\left(-\frac{(k - m)^2}{2 \sigma^2}\right)
\]

to be simple and effective.  Here $\sigma$ is represents the width of the well at $\mathbf{s}^{(m)}$ and is an important parameter.  However, it has been found that in order to maintain numerical stability and also enhance the sequential convergence,   it is better to normalize it. Therfore, the time-dependent weights $ w_k(m) $ in the log-sum-exp term are redefined with explicit normalization:

\[
w_k(m) = \frac{\exp\left(-\frac{(k - m)^2}{2 \sigma^2}\right)}{\sum_{j=0}^{N-1} \exp\left(-\frac{(j - m)^2}{2 \sigma^2}\right)}
\]

This ensures:

\[
\sum_{k=0}^{N-1} w_k(m) = 1 \quad \text{for all } m
\]

The weights $ w_k(m) $ form a Gaussian distribution centered at $ k = m$, assigning the highest influence to the frame closest to the current time step. We note that $w_k(m)$, due to normalization has the structure of a softmax function.\\

Without normalization, the sum $ \sum_k w_k(t) $ varies depending on $ t $, $ \sigma $, and $ N $, due to the finite number of frames and boundary effects (e.g., truncation at $ k = 0 $ or $ k = N-1 $). This variability can cause the log-sum-exp term to fluctuate in magnitude across time steps, leading to inconsistent energy surfaces and unreliable gradients. Normalization guarantees that the weights act as a proper probability distribution over frames, ensuring that the LSE term consistently scales the contributions of $ \langle \mathbf{s}, \mathbf{s}^{(k)} \rangle $. This is particularly important for the weighted softmax probabilities:

\[
p_k(m) = \frac{w_k(m) e^{\beta \langle \mathbf{s}, \mathbf{s}^{(k)} \rangle}}{\sum_j w_j(m) e^{\beta \langle \mathbf{s}, \mathbf{s}^{(j)} \rangle}}
\]

used in the gradient computation (see below), which rely on $ \sum_k w_k(m) = 1 $ to maintain a stable attention mechanism. Furthermore, normalization mitigates numerical instability in the LSE term, especially for large $ \beta $, by preventing the sum from growing excessively.

\subsection{Gradient Descent}

The Gradient descent updates $\mathbf{s}$ as follows:

\begin{equation}
\mathbf{s}_{k+1} = \mathbf{s}_k - \eta \nabla E(\mathbf{s}_k, m)
\label{eqn:descent-2}
\end{equation}

where $\eta$ is the Hebbian learning rate. Taking $\nabla E$, we get: 

\begin{equation}
\nabla E(\mathbf{s}, m) = \lambda \mathbf{s} + 2 \lambda_f \left( \mathbf{s} - \mathbf{s}^{(m)} \right) + 2 \mu \left( \mathbf{s} - \mathbf{s}^{(m-1)} \right) - \sum_{k=0}^{N-1} p_k(m) \mathbf{s}^{(k)} - \mathbf{s}^{(\arg\max_k \langle \mathbf{s}, \mathbf{s}^{(k)} \rangle)}
\end{equation}

where 

\begin{equation}
p_k(m) = \frac{w_k(m) e^{\beta \langle \mathbf{s} , \mathbf{s}^{(k)} \rangle}}{\sum_j w_j(m) e^{\beta \langle \mathbf{s}, \mathbf{s}^{(j)} \rangle}} 
\end{equation}

Our $ p_k(m) $ is a weighted softmax, requiring iterations due to  time shifts, while Ramsauer’s \cite{ramsauer2021hopfield} can converge to a single attractor directly. The fidelity term $ 2 \lambda_f \left( \mathbf{s} - \mathbf{s}^{(m)} \right) $ strongly pulls $ \mathbf{s} $ toward $ \mathbf{s}^{(m)} $, while the continuity term $ 2 \mu \left( \mathbf{s} - \mathbf{s}^{(m-1)} \right) $ ensures the state remains close to the previous retrieved frame, balancing accuracy and smoothness. Normalized weights maintain a consistent LSE contribution, and normalized frames ensure equitable frame competition, reducing the risk of convergence to incorrect minima.\\

These modifications collectively enhance robustness by deepening the target frame’s energy well, smoothing transitions, and stabilizing the optimization landscape, aligning with the goal of reliable, sequential playback as outlined above.\\

\subsection{Fidelity Term}

The fidelity term $ \lambda_f \left\| \mathbf{s} - \mathbf{s}^{m} \right\|^2 $ was introduced in the original functional to penalize deviations of the state $ \mathbf{s} $ from the target frame at time $ t $.  This term contributes to the energy as a quadratic penalty:

\[
\lambda_f \left\| \mathbf{s} - \mathbf{s}^{(m)} \right\|^2 = \lambda_f \sum_{i=1}^d \left( s_i - s_i^{(m)} \right)^2
\]

at $ t = t_m $, where $ s_i $ and $ s_i^{(m)} $ are the $ i $-th components of $ \mathbf{s} $ and $ \mathbf{s}^{(m)} $, respectively.\\

A stronger fidelity term deepens the energy well around $ \mathbf{s}^{(m)} $, ensuring that gradient-based optimization (e.g., steepest descent) converges to the target frame rather than spurious minima. In sequential retrieval, where the initial guess at $ t_{m} $ is $ \mathbf{f}^{(m)} \approx \mathbf{s}^{(m)} $, a large $ \lambda_f $ compensates for potential misalignment by exerting a strong pull toward $ \mathbf{s}^{(m)} $. This is particularly critical when successive frames exhibit significant differences (e.g., scene changes).

\subsection{Continuity Term}

To promote smooth transitions between consecutive frames, we introduce a continuity term:

\[
\mu \left\| \mathbf{s} - \mathbf{s}^{m-1} \right\|^2
\]

At $ t = m $, this becomes:

\[
\mu \left\| \mathbf{s} - \mathbf{s}^{(m-1)} \right\|^2 = \mu \sum_{i=1}^d \left( s_i - s_i^{(m-1)} \right)^2
\]

where $ \mathbf{s}^{(m-1)} \in \mathbb{R}^d $ is the frame retrieved at the previous time step $ {m-1} $, and $ s_i^{(m-1)} $ is its $ i $-th component. For the first frame ($ m = 0 $), we define $ \mathbf{f}^{(-1)} = \mathbf{0} $, making the term $ \mu \|\mathbf{s}\|^2 $, which merges with the regularization term.\\

Movie frames are temporally correlated, with consecutive frames often being visually similar due to small changes in motion or lighting. The continuity term penalizes large jumps between the current state $ \mathbf{s} $ and the previously retrieved frame $ \mathbf{s}^{(m-1)} $, encouraging the optimization to produce $ \mathbf{s}^{(m)} $ that is both close to $ \mathbf{s}^{(m)} $ (via the fidelity term) and consistent with the prior state. This term mimics the smoothness of natural video sequences, reducing abrupt transitions that could occur if the optimization overshoots or converges to an unrelated frame. The parameter $ \mu \geq 0 $ controls the strength of this constraint, with small values (e.g., $ \mu = 0.1 $) typically sufficient to balance continuity and fidelity.

\subsection{Normalization of Frames}

To stabilize the energy functional and ensure consistent dynamics, we normalize the stored frames $ \mathbf{s}^{(k)} $ such that each has a fixed Euclidean norm:

\[
\|\mathbf{s}^{(k)}\| = \sqrt{d}, \quad \text{for } k = 0, 1, \ldots, N-1
\]

where $ d = \text{width} \times \text{height} \times \text{channels} $ is the dimensionality of each frame.  The normalized frames are defined as:

\[
\mathbf{s}^{(k)} \leftarrow \frac{\mathbf{s}^{(k)}}{\|\mathbf{s}^{(k)}\|} \sqrt{d}
\]

assuming $ \|\mathbf{s}^{(k)}\| \neq 0 $. This ensures:

\[
\|\mathbf{s}^{(k)}\|^2 = \sum_{i=1}^d \left( s_i^{(k)} \right)^2 = d
\]

The inner products $ \langle \mathbf{s}, \mathbf{s}^{(k)} \rangle $ in the LSE and max terms drive the alignment of the state $ \mathbf{s} $ with stored frames. If the norms $ \|\mathbf{s}^{(k)}\| $ vary across frames (e.g., due to differences in brightness or contrast), the magnitudes of $ \langle \mathbf{s}, \mathbf{s}^{(k)} \rangle $ become inconsistent, skewing the energy landscape. For instance, a frame with a larger norm could dominate the LSE term regardless of temporal relevance, disrupting sequential retrieval. Normalizing to $ \sqrt{d} $ standardizes the scale, making inner products comparable and ensuring that the weights $ w_k(t) $ primarily determine frame importance. The choice of $ \sqrt{d} $ preserves the expected scale of pixel intensities in high-dimensional frames (e.g., for pixel values in [0, 1], the norm scales appropriately with dimension). This normalization also aids numerical stability in computing $ e^{\beta \langle \mathbf{s}, \mathbf{s}^{(k)} \rangle} $.

\subsection{Mechanism of Time Dependent Gradient Descent}

Using $t_m = m\Delta t$, we can write a simplified form of the energy functional by omitting the fidelity and continuity terms as:

\begin{equation}
E(\mathbf{s}, t_m) = \frac{\lambda}{2} \|\mathbf{s}\|^2 - \frac{1}{\beta} \log \left( \sum_{k=0}^{N-1} w_k(t_m) e^{\beta \langle \mathbf{s}, \mathbf{s}^{(k)} \rangle} \right) -  \max_k \langle \mathbf{s}, \mathbf{s}^{(k)} \rangle
\label{eqn:3}
\end{equation}

This implies that that the energy functional at time $t_m$ is given by the energy surface defined by equation \ref{eqn:3} and the gradient descent will converge to the minimum defined by equation \ref{eqn:3}.  As we transition to time $t_{m+1}$, a new energy surface is defined.  This causes solution trajectory to  ``jump" to the new energy surface and then continue its gradient descent on the new energy surface. (See Appendix 1 for more on jumps).\\

However, this jump is an artifact of the time discretization and the consequent discrete dynamics.  Below we define a ``continuous analog" of equation \ref{eqn:3} above: 

\begin{equation}
E(\mathbf{s}, t) = \frac{\lambda}{2} \|\mathbf{s}\|^2 - \frac{1}{\beta} \log \left( \int_0^T w(t, \tau) e^{\beta \langle \mathbf{s}, \mathbf{s}(\tau) \rangle} d\tau \right) - \max_{\tau} \langle \mathbf{s}, \mathbf{s}(\tau) \rangle
\end{equation}

Note that the summation has been replaced with an integral.  The evolution of the state vector  $\mathbf{s}(t)$ is now a continuous process over $t \in [0, T]$. As $t$ advances continuously, the weight function smoothly shifts weight via the convolution-like integral of $w(t, \tau)$ over $\tau$, causing $ E(\mathbf{s}, t)$ to evolve smoothly. The dynamics of $\mathbf{s}(t)$ follow a gradient flow:

\begin{equation}
\frac{d\mathbf{s}}{dt} = -\nabla E(\mathbf{s}, t)
\end{equation}

So, there are no jumps in the continuous case.  

\subsection{Global Minimum}

In the limit $\sigma \to 0$, where $w_k(m) = \delta_{km}$, it is possible to derive the condition at each time step $m$, the global minimum of $E(s,m)$ is at $s^{(m)}$.  The condition is given by:

\[
\lambda_f > \frac{(2 \lambda_f + 2)^2 \left( \frac{\lambda}{2} + \lambda_f \right)}{(\lambda + 2 \lambda_f)^3}
\]

see Appendix 2 for the derivation.  This represents a transcendental equation in $\lambda_f$. Using,

\[
G(\lambda_f) = \frac{(2 \lambda_f + 2)^2 \left( \frac{\lambda}{2} + \lambda_f \right)}{(\lambda + 2 \lambda_f)^3}
\]

\begin{figure}[h]
\centering
\includegraphics[width=0.75\textwidth]{"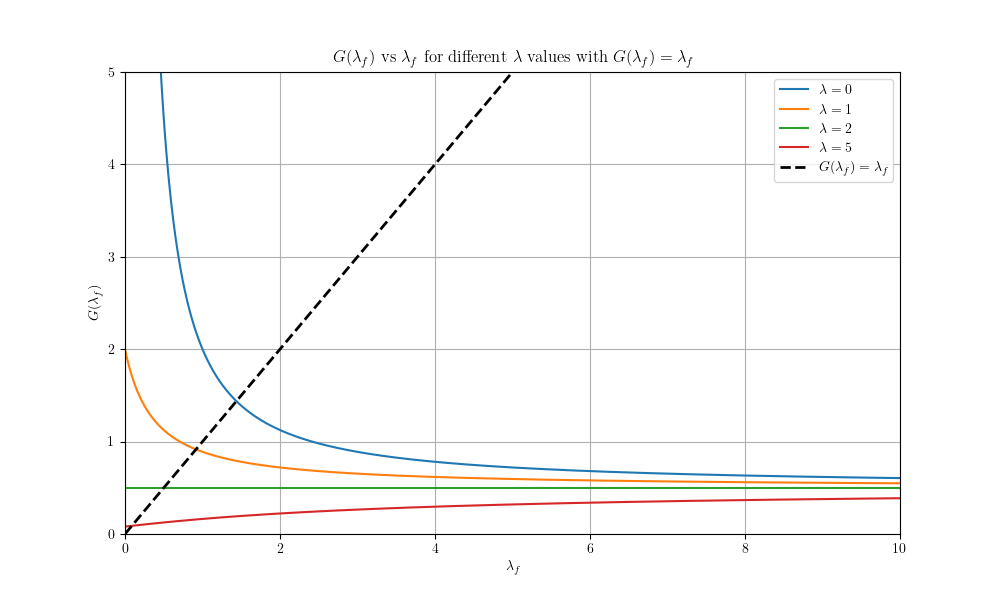"}
\caption{Plot showing $G(\lambda_f)$ vs. $\lambda_f$ for various values of $\lambda$.  The line $G(\lambda_f) =\lambda_f$ is shown as a bold dashed line.}       
\label{plot-1}
\end{figure}

We wish to find the values of $\lambda_f$ for which $\lambda_f > G(\lambda_f)$ for various values of $\lambda$.  Figure \ref{plot-1} shows a plot of $G(\lambda_f)$ vs. $\lambda_f$ for $\lambda \in [0,5]$.  It may be noted that any point to the right of the bold dashed line satisfies the condition $\lambda_f > G(\lambda_f)$. \\

In the next section we define some benchmark movie retrieval problems and conduct numerical experiments to test the storage and retrieval properties of equation \ref{eqn:discrete-enhanced} and investigate the effect of regularization parameters.

\section{Benchmark Problem and Numerical Experiments}

For purposes of testing the energy functional defined by Equation \ref{eqn:discrete-enhanced}  and its dynamics by Equation \ref{eqn:descent-2}, we have utilized six well known cartoon movies from the Blender Foundation, \cite{bigbuckbunny}.  Some parameter of these animated cartoons are summarized in Table \ref{tab:movie_data} which gives the title of each movie, a code name to be utilized as a handy label, the length in seconds, the total number of frames making up the animated movie, and the frames per second (FPS) rate.  The titles of the six animated movies are \quotes{Big Buck Bunny}, abbreviated as BBB, \quotes{Tears of Steel}, abbreviated as TS, and so on.  Some of their properties and meta data are displayed in Table \ref{tab:movie_data}. \\


\begin{table}[h!]
\centering
\caption{Movie Meta Data}
\begin{tabular}{@{}lccccc@{}}
\toprule
\textbf{Title} & \textbf{Code} & \textbf{Total Frames} & \textbf{FPS} & \textbf{Producer} & \textbf{Length} \\
\midrule
Big Buck Bunny                      & BBB   & 19,040   & 30   & Blender Found.   & 634.57 s    \\
Tears of Steel                      & TS    & 17,619   & 24   & Blender Found.   & 734.12 s     \\
Sintel                              & SINT  & 21,313   & 24   & Blender Found.   & 884.04 s    \\
Elephants Dream                     & ED    & 15,691   & 24   & Blender Found.   & 553.79 s     \\
Agent 327: Cloth Simulation         & AGENT & 446      & 24   & Blender Found.   & 18.58 s    \\
\bottomrule
\end{tabular}
\label{tab:movie_data}
\end{table}

In each trial $N$ frames are extracted.  Sometimes these can be the first $N$ frames, or in other cases, the first frame is the $p$th frame, so the frames stored are the frames: $p, p+1, p+2, \ldots p+N-1$. These are stored as vectors $\mathbf{s}^{(0)}, \mathbf{s}^{(1)}, \mathbf{s}^{(2)}, \ldots ,\mathbf{s}^{(N-1)}$.  The first and last vectors, $p$ and $p+N-1$, correspond to time stamps $t_i$ and $t_f$ in the original movie.  The extraction process retrieves vectors $\mathbf{f}^{(0)}, \mathbf{f}^{(1)}, \mathbf{f}^{(2)}, \ldots ,\mathbf{f}^{(N-1)}$, with $\mathbf{f}^{(0)}$ corresponding to $\mathbf{s}^{(0)}$, etc.  If the extraction process is working well then, the extracted vector closely resembles the stored vector, i.e. $\mathbf{f}^{(k)} \approx \mathbf{s}^{(k)}$.  To measure this proximity, we introduce a mean square norm for the $k$th vector, $\text{MSE}_k$ given by: 

\[
\text{MSE}_k =  \frac{1}{d} \| \mathbf{s}^{(k)} - \mathbf{f}^{(k)} \|^2
\]

We also define the retrieval accuracy metric, $\eta$ as:

\[
\eta = \left( \frac{\text{Number of frames with } \text{MSE}_k < 0.05}{N} \right) \times 100
\]

The results of the trials are given in Table \ref{tab:trials} where the movie clips were imported as frames of size $d = 256 \times 256 \times 3 = 196608$.  The parameters $\beta$, $\lambda$, $\lambda_f$, $\sigma$, $\mu$ in the energy functional given by Equation \ref{eqn:discrete-enhanced} were carefully tuned to increase the accuracy, $\eta$. \\

\begin{table}[h!]
\fontsize{8pt}{8pt}\selectfont 
\centering
\caption{List of Trials}
\begin{tabular}{@{}ccccccccccccccc@{}}
\toprule
\text{Trial \#} & \text{Clip}  & p & $t_i$ & $t_f$ & $N$ & $d$ & $\beta$  & $\sigma$    &  $\lambda$ & $\lambda_f$ & $\mu$ &$\eta$ & $\epsilon$ & $S$\\
\midrule
{ 1} & BBB    & 1    & 0.03  & 13.33  & 400 &196608  & 1   & 2 &   0.01  & 500 & 0.001 & 100\% & 6 & 2 \\
{ 2} & BBB    & 125  & 4.17  & 4.33   & 6   &196608  & 1   & 2 &   0.01  & 500 & 0.001 & 100\% & 5 & 2 \\
{ 3} & BBB    & 1000 & 33.33 & 53.29  & 600 &196608  & 1   & 2 &   0.01  & 500 & 0.001 & 100\% & 0 & 3 \\
{ 4} & BBB    & 1500 & 49.99 & 76.62  & 800 &196608  & 1   & 2 &   0.01  & 500 & 0.001 & 100\% & 1 & 6 \\
{ 5} & BBB    & 2300 & 76.65 & 109.95 &1000 &196608  & 1   & 2 &   0.01  & 500 & 0.001 & 100\% & 1 & 8 \\
{ 6} & BBB    & 3300 & 109.98& 176.61 &2000 &196608  & 1   & 2 &   0.01  & 500 & 0.001 & 100\% & 2 & 15 \\
{ 7} & AGENT  & 1    & 0.04  & 16.17  & 400 &196608  & 1   & 2 &   0.01  & 500 & 0.001 & 100\% & 0 & 6 \\
{ 8} & TS     & 200  & 8.33  & 29.12  & 500 &196608  & 1   & 2 &   0.01  & 500 & 0.001 & 100\% &13 & 4 \\
{ 9} & BBB    & 100  & 0.03  & 13.33  & 400 &196608  & 1   & 2 &   0.01  & 500 & 0.001 & 100\% & 6 & 2 \\
\bottomrule
\end{tabular}
\label{tab:trials}
\end{table}

The \texttt{python} binding of the OpenCV library \texttt{cv2}  was used for reading/writing and formatting of the MP4 clips. 
The gradient descent was accomplished by using the \texttt{minimize} function from the \texttt{python} library \texttt{scipy} with the \texttt{method} set to \texttt{L-BFGS-B} and \texttt{tol} set to $1 \times 10^{-5}$.  The energy function \texttt{fun} and the gradient function \texttt{jac} were defined as python functions and passed in as arguments.  The initial guess \texttt{x0} was set to $\mathbf{s}^{(m-1)}$. All vector operations were performed using vectorized \texttt{numpy} function calls. It may be noted that each gradient step scales as $O(Nd)$ and total steps scale as $O(N^2d)$. \\

\begin{figure}[h]
     \centering
     \includegraphics[width=1.00\textwidth]{"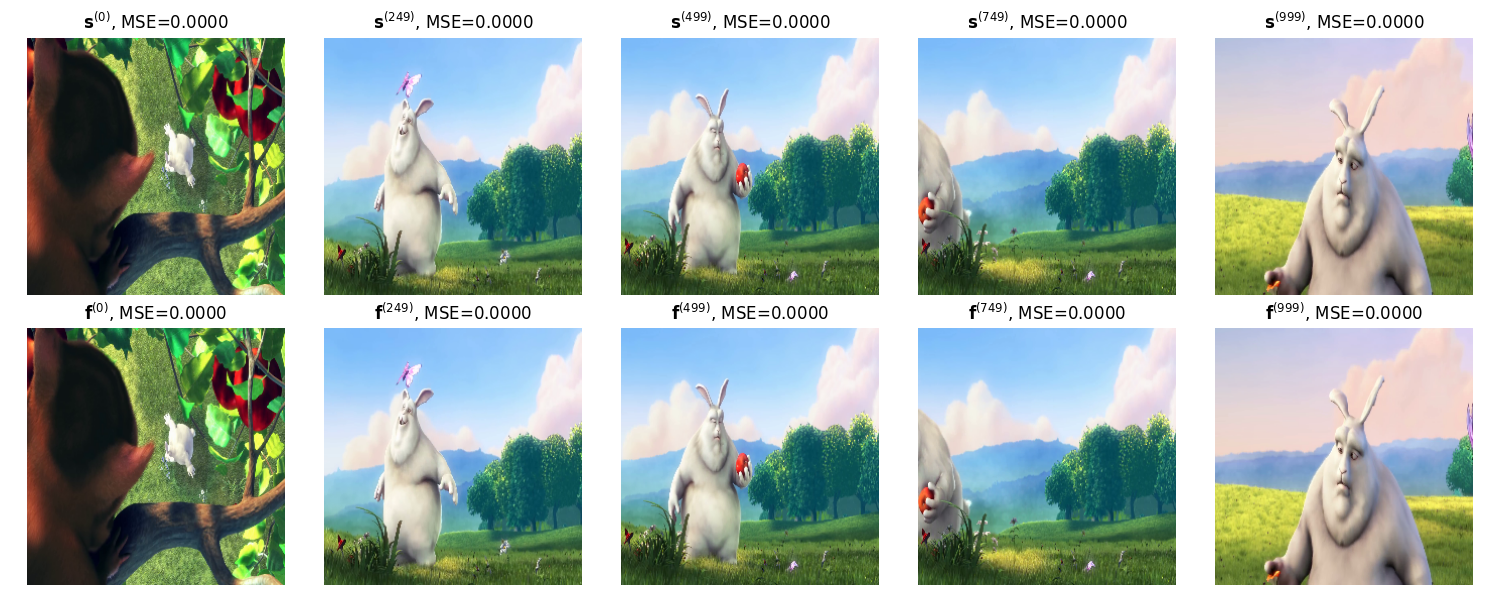"}         
     \caption{Original frames $\mathbf{s}^{(0)}$, $\mathbf{s}^{(249)}$, $\mathbf{s}^{(499)}$,$\mathbf{s}^{(749)}$ and retrieved frames $\mathbf{s}^{(999)}$ and the retrieved frames $\mathbf{f}^{(0)}$,  $\mathbf{f}^{(249)}$, $\mathbf{f}^{(499)}$,$\mathbf{f}^{(749)}$, $\mathbf{f}^{(999)}$ for Trial $\# 3$. }       
     \label{frames-1}
\end{figure}

Trials $\#1$ - $\#6$ of the movie BBB were performed by varying the starting frame $p$ and the length of the clip $N$.  For instance trial $\# 1$ started at $p=1$ and included $N=400$ frames, going from $t_i=0.1$ to $t_f =3$ with a frame retrieval accuracy of $\eta =100\%$.  The retrieved frames were stored as an MP4 file and played and inspected and to the eye they appeared to play much like the original, without frames appearing out of order.  The last column in Table \ref{tab:trials}  gives, $S$, the number of \quotes{scene changes}.  These are important because at onset of a new scene, the frame vector could potentially shift far away from the previous frame, and thus increase the chance of not the steepest descent being thrown off.  However, from playing the movie and the fact that $\text{MSE}_K$ was much smaller than the threshold, is an indication of the robustness of the algorithm.   A sample of five original frames $\mathbf{s}^{(0)}$, $\mathbf{s}^{(249)}$, $\mathbf{s}^{(499)}$,$\mathbf{s}^{(749)}$ and $\mathbf{s}^{(999)}$ and the retrieved frames $\mathbf{f}^{(0)}$,  $\mathbf{f}^{(249)}$, $\mathbf{f}^{(499)}$,$\mathbf{f}^{(749)}$, $\mathbf{f}^{(999)}$ have been shown in Figure \ref{frames-1} corresponding to Trial $\# 3$..  As can be seen the $\text{MSE}_k$ for each frame is less than $1\times 10^{-4}$.  The frames are indistinguishable to the eye in terms of color, tone, etc. \\

It is notable that scaling $N$ from $400$ to $2000$ did not degrade the storage/retrieval process accuracy because as we noted earlier, with sufficiently high $\lambda_f$ and low $\sigma$, the next frame is always the global minimum.  In principle therefore, there is no limitation to, for example, storing a 20 minute, $\text{FPS}=30$ clip (with $N = 30\times60\times 20 = 36,000$) or even a 2-hour ($N = 30 \times 60 \times 120 = 216,000$)  movie in this manner, however the time required for retrieval will increase as $N^2$.\\

\newpage

\begin{figure}[h]
     \centering
     \includegraphics[width=1.00\textwidth]{"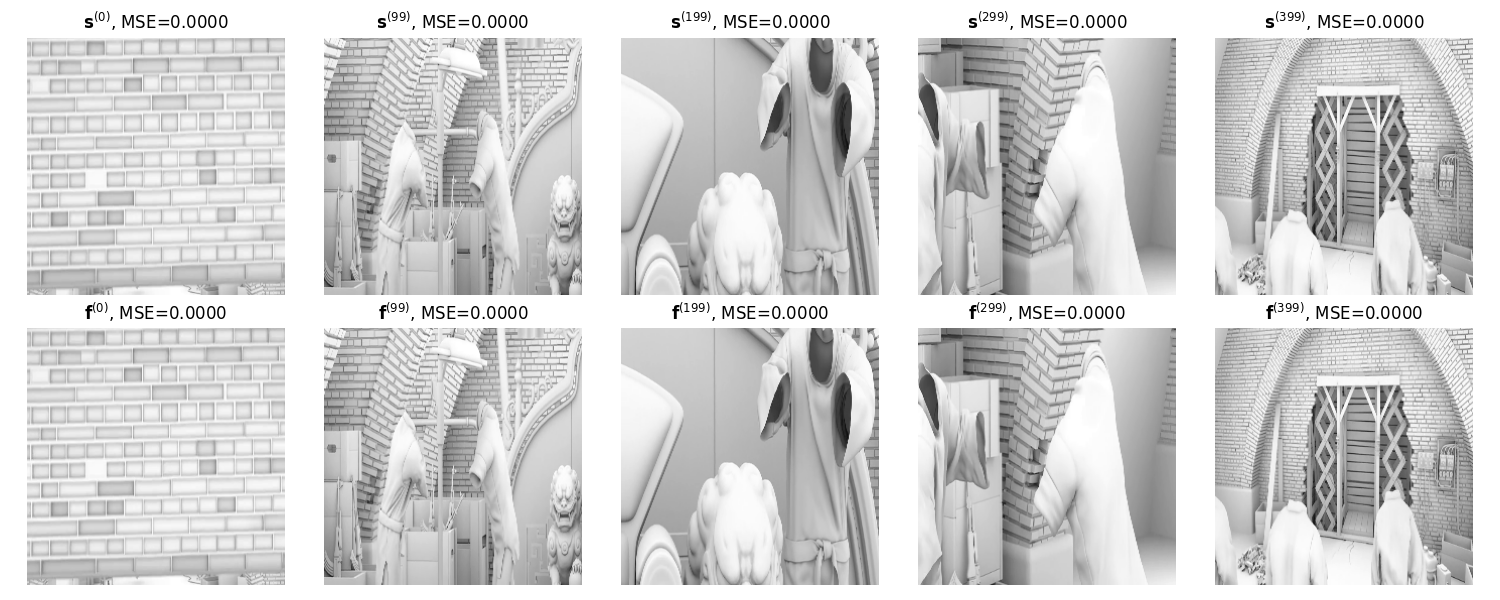"}         
     \caption{Original frames $\mathbf{s}^{(0)}$, $\mathbf{s}^{(99)}$, $\mathbf{s}^{(199)}$,$\mathbf{s}^{(299)}$ and  $\mathbf{s}^{(399)}$ and the retrieved frames $\mathbf{f}^{(0)}$,  $\mathbf{f}^{(99)}$, $\mathbf{f}^{(199)}$, $\mathbf{f}^{(299)}$, $\mathbf{f}^{(399)}$ for Trial $\# 7$. }       
     \label{frames-2}
\end{figure}

Trial $\# 7$ stores and retrieves $N=400$ frames from the clip AGENT, again with $100\%$ accuracy. The orginal frames and retrieved frames at five equally spaced intervals are shown in Figure \ref{frames-2}, and again it can be see that the retrieved images are faithful to the original.\\

\section{Conclusions and Suggestions for Future Work}

We have developed a time dependent energy functional which allows the creation of Hopfield energy surface that prioritizes the convergence to the local frame at each time step.  We have then applied this to movie frame retrieval and tested tested several well known animated clips of varying length.  The results demonstrate that this approach leads to high accuracy in sequential frame retrieval.\\

This approach also opens several promising avenues for future investigation, particularly in validating its efficacy across diverse sequential long-memory tasks.  \\

For example the Long-Range Arena (LRA) tasks proposed by Tay et al. \cite{tay2020long} offer a rigorous testbed for long-range dependencies, with sequences up to 4,096 tokens across domains like text classification and pathfinding, allowing an assessment of how the Gaussian kernel enhances attention over extended contexts compared to vanilla transformers. \\

Similarly, character-level language modeling on the Penn Treebank dataset can explore the model’s ability to maintain linguistic coherence over thousands of characters, with perplexity as a key metric, building on the memory-augmented approaches of Sukhbaatar et al. \cite{sukhbaatar2015end}\\

Time-series forecasting, using datasets like hourly electricity load, presents another opportunity to leverage the temporal weighting of the Gaussian kernel for capturing seasonal patterns, with evaluation through Mean Absolute Error (MAE) to compare against traditional time-series models. The copying memory task, a synthetic benchmark from Graves et al. \cite{graves2014neural}, can test the model’s precision in retrieving random sequences after delays, offering insights into memory retention and forgetting dynamics influenced by the regularization parameter $ \lambda $.\\

Further exploration could include the bAbI question answering tasks from Weston et al. \cite{weston2015towards}, where the model’s ability to retrieve and reason over multiple supporting facts in long stories can be assessed through accuracy metrics, potentially integrating the log-sum-exp functional for smooth memory aggregation.\\

\section*{Appendix 1: Mechanism of Time Dependent Gradient Descent}

In the context of equation \ref{eqn:3}, the energy functional is a function both of the the state space $\mathbf{s}$ and time $t$, $E = E(\mathbf{s},t)$.  This complicates the overall picture of getting to steepest descents, because now the whole energy surface shifts and morphs in time. Let us explore the structure and mathematical properties of this time-dependent energy functional and see how gradient descent operates on this surface.  Let us first consider the energy surface at any time $t=t_m$ given by:\\

\begin{equation}
E(\mathbf{s}, t_m) = \frac{\lambda}{2} \|\mathbf{s}\|^2 - \frac{1}{\beta} \log \left( \sum_{k=0}^{N-1} w_k(t_m) e^{\beta \langle \mathbf{s}, \mathbf{s}^{(k)} \rangle} \right) -  \max_k \langle \mathbf{s}, \mathbf{s}^{(k)} \rangle
\label{eqn:3}
\end{equation}

From the above equation it is immediately apparant that the overall structure of the energy surface at a fixed time, arises from the interaction between its three constitutive terms:  

\begin{enumerate}[label=(\roman*)]
 \item The Regularization term, $ \frac{\lambda}{2} \|\mathbf{s}\|^2 $, which is quadratic in $\mathbf{s}$, forming a paraboloid centered at $\mathbf{s} = 0$. It contributes a convex, upward-curving component, ensuring $ E \to +\infty$ as $\|\mathbf{s}\| \to \infty$.
 
 \item The LSE term, $-\frac{1}{\beta} \log \left( \sum_k w_k(t) e^{\beta \langle \mathbf{s}, \mathbf{s}^{(k)} \rangle} \right)$ which is a Log-sum-exp and approximates a soft maximum.  As $\beta \to \infty $, it approaches $ -\max_k \langle \mathbf{s}, \mathbf{s}^{(k)} \rangle $, but is weighted by $ w_k(t)$. For high $\beta$, it is nearly linear near each $\mathbf{s}^{(k)}$, but smooths into a concave dip where $w_k(t)$ is large.
 
 \item The Max term, $ -\max_k \langle \mathbf{s}, \mathbf{s}^{(k)} \rangle $, inspired by a similar term in the well known study by Vaswani et al. \cite{vaswani2017attention} is a piecewise linear term, subtracting the strongest similarity, deepening the bowl where $\mathbf{s}$ aligns with a frame.
\end{enumerate}

At $t_k=0$, and with $\sigma = \frac{1}{2}$, the weight function may be written as $w_m =  \exp\left[-\left(k \right)^2\right] $, giving for the energy:

\begin{align}
E(\mathbf{s}, 0) &= \frac{\lambda}{2} \|\mathbf{s}\|^2 - \frac{1}{\beta} \log \left[  e^{0}e^{\beta \langle \mathbf{s}, \mathbf{s}^{(0)} \rangle} + e^{-1}e^{\beta \langle \mathbf{s}, \mathbf{s}^{(1)} \rangle} + e^{-4}e^{\beta \langle \mathbf{s}, \mathbf{s}^{(2)} \rangle} \right. \nonumber\\
&\left. + \; \ldots \; + e^{-2(N-1)}e^{\beta \langle \mathbf{s}, \mathbf{s}^{(N-1)} \rangle}\right] -  \max_k \langle \mathbf{s}, \mathbf{s}^{(k)} \rangle
\label{eqn:4}
\end{align}

Several things may be noted from equation \eqref{eqn:4}: (i) In the LSE term, contributions from frames that are further away in time from $\mathbf{s}^{(0)}$ fall off very quickly in strength.  So the weights, $w_k(0)$, act has dampeners for all the other minima, except $\mathbf{s^{(0)}}$.  In the present case, the term with the smallest penalty due to $w_k(0)$ is the term $e^{\beta \langle s,s^{(0)}\rangle)}$, which is $1$, and the next term $e^{-1}e^{\beta \langle s,s^{(1)}\rangle)}$, the  penalty is $e^{-1}$.  For other terms, it is higher powers of inverse $e$.  (ii)  as expected, $\beta$ plays a very important role in amplifying the alignment between $\mathbf{s}$ and $\mathbf{s}^{(0)}$  (iii) although we have assumed $\sigma=\frac{1}{2}$, in the above equation, $\sigma$ provides an important scaling factor for the weight function $w_n(t)$. For smaller $\sigma$, we will see a little later that the weight functon effectively becomes a delta function and in this limit, it is possible to prove many useful results. (iv) In terms of dynamics, the steepest descent process drives the convergence to $\mathbf{s}^{(0)}$.\\

In the next step, at $t=t_2$, the energy functional $E(\mathbf{s},2)$ is given by:

\begin{align}
E(\mathbf{s}, 1) &= \frac{\lambda}{2} \|\mathbf{s}\|^2 - \frac{1}{\beta} \log \left[  e^{-1}e^{\beta \langle \mathbf{s}, \mathbf{s}^{(0)} \rangle} + e^{0}e^{\beta \langle \mathbf{s}, \mathbf{s}^{(1)} \rangle} + e^{-1}e^{\beta \langle \mathbf{s}, \mathbf{s}^{(2)} \rangle} \right. \nonumber \\
&+ \left. \; \ldots \; + e^{-2(N-2)}e^{\beta \langle \mathbf{s}, \mathbf{s}^{(N-1)} \rangle} \right] -  \max_k \langle \mathbf{s}, \mathbf{s}^{(k)} \rangle
\end{align}

So at $t=1$, the energy surface, $E(\mathbf{s},1)$ has shifted.  With $\sigma =\frac{1}{2}$, the weight function is now given by $w_k(1) = \exp\left[-(k - 1 )^2 \right] $ and the result of the shift is that the term with the least penalty of $1$ is $e^{0}e^{\beta \langle s,s^{(1)}\rangle)}$, whereas the terms $e^{-1}e^{\beta \langle s,s^{(0)}\rangle)}$ and $e^{-1}e^{\beta \langle s,s^{(2)}\rangle)}$ have the penalty $e^{-1}$.  All other terms have higher penalties.  Thus the minimum energy well around $\mathbf{s^{(1)}}$ is going to be the dominant one and the steepest descent process is going to lead to $\mathbf{s^{(1)}}$.\\

One of the notable features of the dynamics is that the trajectory of the solution has to \quotes{jump} to a new surface at the end of the current time period.  In any given interval, the progression of the steepest descents is smooth and occurs over the current energy surface.  But with the onset of a new time interval, the energy surface itself changes and the trajectory experiences a jump and then proceeds smoothly to the new minimum.  As we shall see shortly, these jumps are the result of a descrete time advancement process.  If we move to a continuous time advancement, then there would be a single smooth trajectory that would govern the evolution. \\

\begin{figure}[h]
        \centering
        \subfloat[$t=0$]{%
            \includegraphics[width=0.50\textwidth]{"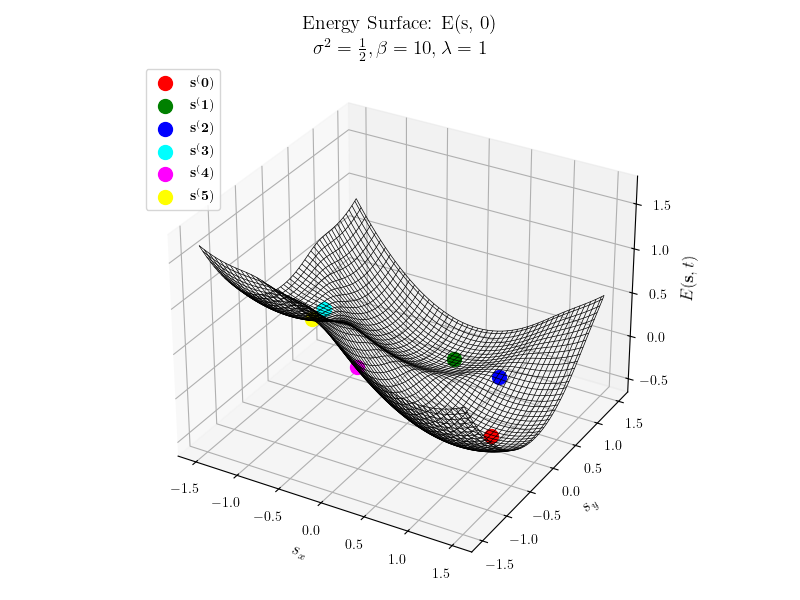"}\label{v1}}
        \subfloat[$t=1$]{%
            \includegraphics[width=0.50\textwidth]{"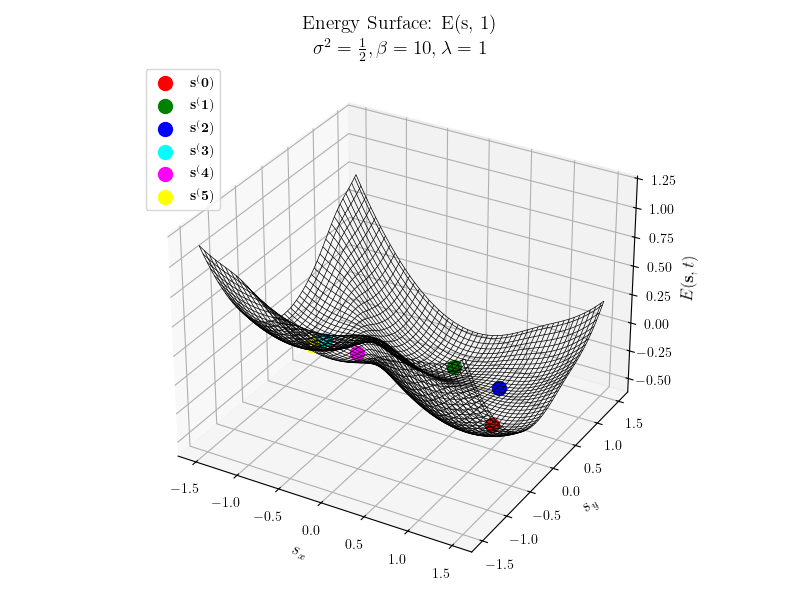"}\label{v2}}\\
        \subfloat[$t=2$]{%
            \includegraphics[width=0.50\textwidth]{"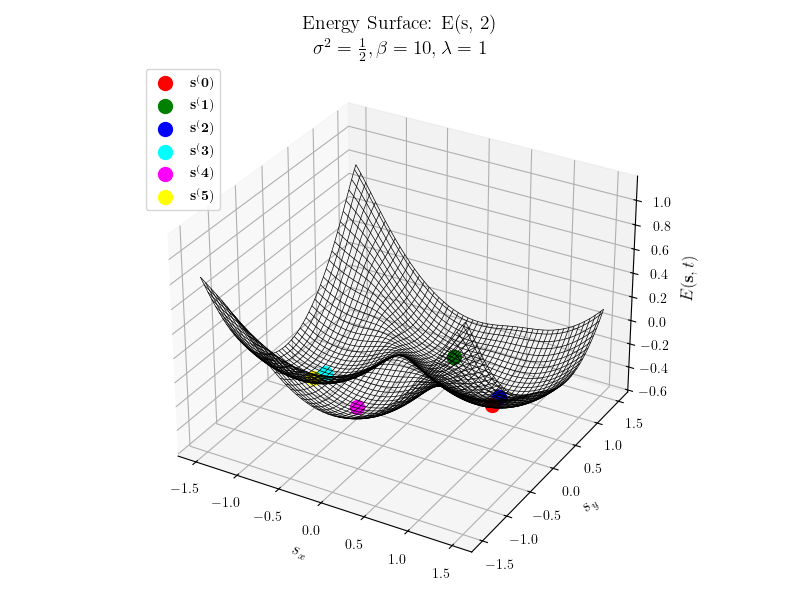"}\label{v1}}
        \subfloat[$t=3$]{%
            \includegraphics[width=0.50\textwidth]{"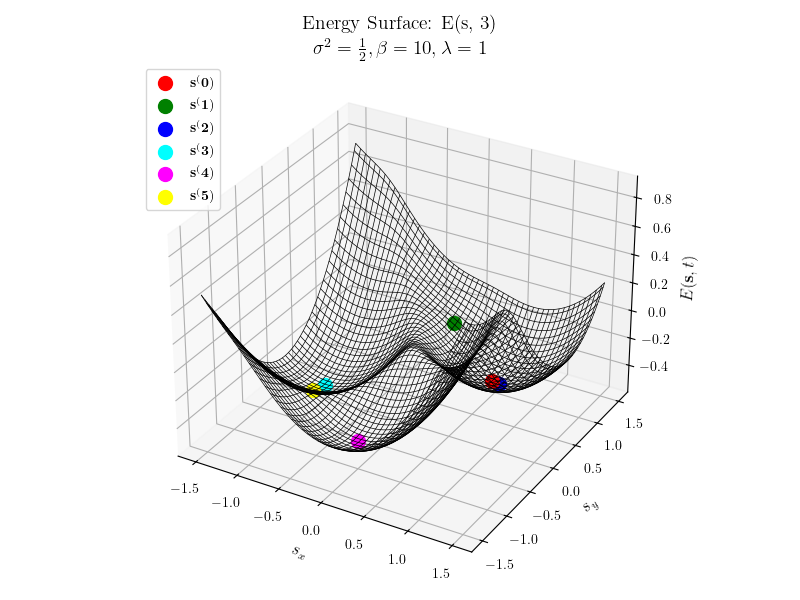"}\label{v2}}\\    
            \caption{Energy Surface showing minima and barriers for (a) $t=0$, (b) $t=1$, (c) $t=2$, (d) $t=3$ and the shifting attention.}       
        \label{barriers-1}
\end{figure}

To visualize the energy surfaces as they morph at each time step, consider the following setup:  $5$ vectors of dimension $2$ (that is, we have $N=5$ and $d=2$)  given by $\mathbf{s}^{(0)} = [1, 0]$, $\mathbf{s}^{(1)} = [0, 1]$, $\mathbf{s}^{(2)} = [1, 1]$, $\mathbf{s}^{(3)} = [-1, 0]$, $\mathbf{s}^{(4)} = [0, -1]$, $\mathbf{s}^{(5)} = [-1, -1]$.   We set $\lambda =1$, $\beta =100$ and $\sigma^2=0.5$ and plot the the energy surfaces $E(\mathbf{s},0)$, $E(\mathbf{s},1)$, $E(\mathbf{s},2)$ and $E(\mathbf{s},3)$ etc. in Figure \ref{barriers-1}(a-d) to show the minima and the barriers in the energy landscape.  Looking at Figure \ref{barriers-1}(a), we can see that while there are several minima, the minimum for $\mathbf{s}^{(0)} = [1, 0]$ is one of the prominent ones.   It is not possible to tell just by inspection that this is the deepest minimum, however it is one of the deepest ones.   However, as we transition from $E(\mathbf{s},0)$ to $E(\mathbf{s},1)$, the most prominent minimum well shifts from $\mathbf{s}^{(0)}$ to $\mathbf{s}^{(1)}$ as can be seen in Figure \ref{barriers-1}(b). In the next section, we will show that the solution is indeed guaranteed to advance to the next frame. In Figure \ref{barriers-1}(c) we can see the shift of the dynamic attention from $\mathbf{s}^{(1)}$ to $\mathbf{s}^{(2)}$ with the deepest well at $\mathbf{s^{(2)}}$.  Similar shift in the energy surface can be seen in Figure \ref{barriers-1}(d). \\

Another important point to be noted is that there is no continuous path on one a single surface since $E(\mathbf{s}, t)$  is a family of surfaces, not a single landscape.  Transition isn’t a gradient descent across $t$; it is a jump to a new surface $ E(\mathbf{s}, t_1) $, followed by descent within that bowl. The energy jump between $E(\mathbf{s}^{(0)}, t_0) < E(\mathbf{s}^{(0)}, t_1) $  is necessary  as the old minimum is no longer optimal.  This jumps is necessiatated by the discrete time intervals.\\

\section*{Appendix 2: Summary of the Global Minimum Condition}

We consider the case of the limit $\sigma \to 0$, since this offers a simplified energy functional.  Then, using $w_k(m) = \delta_{mk}$ gives the energy functional as:

\[
E(\mathbf{s}, m) = \frac{\lambda}{2} \|\mathbf{s}\|^2 + \lambda_f \|\mathbf{s} - \mathbf{s}^{(m)}\|^2 + \mu \|\mathbf{s} - \mathbf{s}^{(m-1)}\|^2 - \langle \mathbf{s}, \mathbf{s}^{(m)} \rangle - \max_k \langle \mathbf{s}, \mathbf{s}^{(k)} \rangle
\]

This functional is minimized at $\mathbf{s}^{(m)}$ at step $m$. The use of delta function weights simplifies the non-convex Log-Sum-Exponential form, making the optimization more tractable.\\

To confirm that $\mathbf{s}^{(m)}$ is the global minimum, we require $E(\mathbf{s}^{(m)}, m) < E(\mathbf{s}, m)$ for all $\mathbf{s} \neq \mathbf{s}^{(m)}$. Evaluating the functional at $\mathbf{s} = \mathbf{s}^{(m)}$ yields:

\[
E(\mathbf{s}^{(m)}, m) = \frac{\lambda d}{2} + 2 \mu d - 2d
\]

Here, we assume $\|\mathbf{s}^{(m)}\|^2 = d$ and $\langle \mathbf{s}^{(m)}, \mathbf{s}^{(m-1)} \rangle \approx 2d$, reflecting the geometric properties of the iterates.\\

For an arbitrary $\mathbf{s} \in [0,1]^d$, we derive a lower bound for the energy:

\[
E(\mathbf{s}, m) \geq \left( \frac{\lambda}{2} + \lambda_f \right) \|\mathbf{s}\|^2 - (2\lambda_f + 2) \|\mathbf{s}\| \sqrt{d} + \lambda_f d
\]

The energy difference is defined as $\Delta E = E(\mathbf{s}, m) - E(\mathbf{s}^{(m)}, m)$. To analyze this, we introduce a quadratic function:

\[
f(t) = \left( \frac{\lambda}{2} + \lambda_f \right) t^2 - (2 \lambda_f + 2) \sqrt{d} \, t
\]

where $t = \|\mathbf{s}\|$. This function is minimized at:

\[
t_0 = \frac{(2 \lambda_f + 2) \sqrt{d}}{\lambda + 2 \lambda_f}
\]

The energy difference satisfies:

\[
\Delta E \geq f(t_0) + \lambda_f d + 2d - \frac{\lambda d}{2} - 2 \mu d
\]

For $\Delta E > 0$, we require:

\[
\lambda_f > \frac{(2 \lambda_f + 2)^2 \left( \frac{\lambda}{2} + \lambda_f \right)}{(\lambda + 2 \lambda_f)^2} + \frac{\lambda}{2} - 2 + 2 \mu
\]

In the bounded domain ($\mathbf{s} \in [0,1]^d$), the fidelity term dominates, leading to:

\[
\Delta E \geq \lambda_f \|\mathbf{s} - \mathbf{s}^{(m)}\|^2 + f(t_0)
\]

Given that $\left\| \mathbf{s} - \mathbf{s}^{(m)} \right\|^2 \leq d$, the condition simplifies to:

\[
\lambda_f > \frac{(2 \lambda_f + 2)^2 \left( \frac{\lambda}{2} + \lambda_f \right)}{(\lambda + 2 \lambda_f)^2}
\]

The final condition for $\mathbf{s}^{(m)}$ to be the global minimum is:

\[
\lambda_f > \frac{(2 \lambda_f + 2)^2 \left( \frac{\lambda}{2} + \lambda_f \right)}{(\lambda + 2 \lambda_f)^2}
\]

This transcendental inequality, solvable numerically for specific $\lambda$, ensures that the fidelity term creates a deep energy valley at $\mathbf{s}^{(m)}$, securing its position as the global minimum.

\newpage
\bibliography{BlockBibliography}
\bibliographystyle{amsplain}

\end{document}